# A hybrid spatial data mining approach based on fuzzy topological relations and MOSES evolutionary algorithm

Amir Hossein Goudarzi
Department of Electrical and Computer Engineering,
Isfahan University of Technology
ah.goudarzi@ec.iut.ac.ir

Nasser Ghadiri[1]
Department of Electrical and Computer Engineering,
Isfahan University of Technology
nghadiri@cc.iut.ac.ir

**Abstract:** Making high-quality decisions in strategic spatial planning is heavily dependent on extracting knowledge from vast amounts of data. Although many decision-making problems like developing urban areas require such perception and reasoning, existing methods in this field usually neglect the deep knowledge mined from geographic databases and are based on pure statistical methods. Due to the large volume of data gathered in spatial databases, and the uncertainty of spatial objects, mining association rules for high-level knowledge representation is a challenging task. Few algorithms manage geographical and non-geographical data using topological relations. In this paper, a novel approach for spatial data mining based on the MOSES evolutionary framework is presented which improves the classic genetic programming approach. A hybrid architecture called GGeo is proposed to apply the MOSES mining rules considering fuzzy topological relations from spatial data. The uncertainty and fuzziness aspects are addressed using an enriched model of topological relations by fuzzy region connection calculus. Moreover, to overcome the problem of time-consuming fuzzy topological relationships calculations, this a novel data pre-processing method is offered. GGeo analyses and learns from geographical and non-geographical data and uses topological and distance parameters, and returns a series of arithmetic-spatial formulas as classification rules. The proposed approach is resistant to noisy data, and all its stages run in parallel to increase speed. This approach may be used in different spatial data classification problems as well as representing an appropriate method of data analysis and economic policy making.

---

[1] *Corresponding author. Address: Department of Electrical and Computer Engineering, Isfahan University of Technology, Isfahan, Iran. Phone : +98-31-3391-9058, Fax: +98-31-3391-2450, Alt. email: nghadiri@gmail.com*

**1. Introduction**

Massive amounts of spatial data have been used in many fields in various applications such as urban planning, transport, telecommunications, etc. These data are stored in the Geographic Database Management Systems (GDBMS) and are handled and analyzed by the Geographic Information Systems (GIS). This large amount of spatial data contains knowledge that cannot be discovered using GIS techniques. Intelligent approaches like Knowledge Discovery in Databases (KDD) have been offered to tackle such problems. Different solutions have been proposed for KDD in geographic databases. Most methods use query languages or new spatial data mining algorithms. The problem is that geographic databases are based on querying through Structured Query Language (SQL) and do not implement data mining languages. Neither do they perform geographic data preprocessing, that seriously affects the quality of the analysis process.

A particular kind of KDD called Geographical Knowledge Discovery (GKD) is introduced to extract knowledge from geographic databases. In recent years, geospatial data have experienced information "explosion"; this means trespassing from an era of small amounts of data to the world of big data. This data contains features which challenge old solutions. Most data domains used in KDD are multi-dimensional but independent; so traditional data mining methods seem useless. Spatial data have many dimensions as well as four highly related information dimensions and represent a measuring framework for other dimensions. This framework affects geographical attributes and the extracted patterns. One of the most popular frameworks is the one consistent with the Euclidean geometry and distance model. Other frameworks are used in different contexts. Also, geographic attributes are usually spatially dependent; which means that these attributes tend to happen in particular locations. These places are proximal in the Euclid space. Even though, direction, connection and other geographical properties (like vegetation) may affect the spatial dependency. On the other hand, in high dimensional data compared to low-dimensional data, relations like distance, direction, and connection are more complicated to define and calculate. Also, different geographical digital data types bring up lots of more challenges.

Classification rule mining is one of the most important approaches for knowledge discovery in databases. Different techniques and algorithms for extracting the classification rules have been extensively studied because of the variety of their practical applications. Mining classification rules usually uses supervised learning methods that are to discover patterns of training data so that the resulting rules can be applied in the classification of other datasets as well.

It has been shown that evolutionary methods like genetic programming can be used efficiently to classify geographic data(Freitas, 2002,Parpinelli et al., 2002). Meta-optimizing semantic evolutionary search (Looks, 2007) (MOSES) is a new evolutionary approach that stems from genetic programming and has been successfully applied to solve hard problems in many application with accurate results.

In this paper, a novel data mining architecture called GGeo is proposed to efficiently mine spatial data and facilitate knowledge discovery in geographic databases. Moreover, we propose an efficient preprocessing method implemented using Fuzzy Regional Connection Calculus (RCC) to prepare geographic data for the GGeo or other models. The proposed methods are evaluated in a real-world problem to add an urban growth parameter to the classic road planning decision problem.

The paper is organized as follows. An overview of the related work is given in Section 2. Section 3 describes our proposed GGeo architecture for evolutionary mining of spatial data, after a brief outline of the basic concepts. The experimental results are presented in Section 4. Section 5 concludes the paper and points out to future research.

## 2. Related Work

Geographic databases store spatial features which represent real world entities. Spatial features are part of a feature type and may have additional non-spatial attributes. In relational databases, every feature type is usually stored in a different table or relation (Shekhar and Chawla, 2003). In data mining through database query languages, Han et al. presented GMQL which was introduced in GeoMiner Software(Han et al., 1997). Later, Apice et al. showed a geographical properties extractor named ARES which calculates all Boolean spatial relations between all spatial feature types, which limits at applicability for big data (Appice et al., 2005). This approach exploited the Spada algorithm (Lisi and Malerba, 2002). Apice et al. later presented a unified data mining framework in GIS environments (Appice et al., 2007). In general, the classification methods for spatial data, build the model based on a set of relevant attributes and their values and subsequently map data instances to pre-defined classes.

Ester, et al. introduced one of the earliest non-query approaches. They used a machine learning algorithm based on ID3 to create a set of spatial classification rules based on the attributes of a spatial entity and the degree of dependency on its neighbors. The user provides a maximum search length for evaluating the relationships between instances. Adding a rule to the tree requires satisfying a minimum threshold of information gain (Ester et al., 1997). Sitanggang et al. have also used an extended similar approach (Sitanggang et al., 2011). In spatial data mining using coordinates transformation, Korting et al. presented an approach for analyzing spatiotemporal data. Their GeoDMA method creates a set of descriptive features that improve the classification accuracy (Körting et al., 2013).

In some studies, several classification algorithms have been integrated into a unified model. Liu et al. presented a C++ framework which incorporates a range of classification algorithms to predict the geographical distribution of a particular event. As a more particular domain, spatial data mining approaches are used to manage urban planning problems. Spatial data mining was also used for finding a relationship in practical cases like oil extraction (Cai et al., 2014). However, the large volume of spatial data makes the accuracy of these methods limited for some real-world applications.

Evolutionary algorithms have been thoroughly studied to overcome this problem. In this area, Genetic Programming (GP) is exploited as a method to automatically generate computer programs using a process similar to biological evolution (Koza, 1992). GP has been applied to a range of data mining tasks including feature selection and classification of data (Tran et al., 2016) . One advantage of using

GP for classification tasks is that by allowing the training time to increase, the results get more accurate until the point where over-fitting occurs. So if sufficient time is available to train the classifiers, a better genetic classifier will be produced by the algorithm that potentially out-performs other classification methods (e.g. C4.5) (Loveard and Ciesielski, 2001). GP is highly probabilistic and different runs lead to different results. This variability of solutions gives GP a voting strategy to produce more accurate classification results (Loveard and Ciesielski, 2001). Our proposed method is based on a novel approach from the GP category named MOSES that will be introduced later in this section. MOSES could be categorized as a hybrid hill climbing method (Kim et al., 2014) and is recently used in other application domains including healthcare systems (Poulin et al., 2016).

Another problem with the classification methods for spatial data mining is neglecting the tremendous amount of valuable knowledge that can be extracted from geographical databases by spatial relationships. The spatial attributes of geographic object types have intrinsic spatial relationships (e.g. *close*, *far*, *contains* and so on). Spatial relationships are not explicitly stored in a database, and they should be calculated with spatial operations (Bogorny et al., 2005). The three main spatial relationships are distance, cardinal relationships, and topological relations. A common formalization of topological relationships is Region Connection Calculus (RCC) (Bogorny et al., 2006). For preparing the geographic databases and preprocessing of data for data mining using RCC, a primary feature type is selected and the spatial relations between this feature and relevant ones will be computed. The main feature and related features are stored in different, independent database tables. The result of this data preparation will be represented in a file by a row for each instance, having several columns that contain non-spatial attributes and one or more columns with appropriate feature types (e.g., *contains* river). Given a set of instances $M = \{m_1, m_2, \ldots, m_n\}$ which is the main feature type, $R = \{O_1, O_2, \ldots, O_m\}$ which is the appropriate feature type set and $O_i = \{o_1, o_2, \ldots, o_p\}$ is a relevant feature (e.g., river set which contains different rivers). The extraction of spatial relationships requires every instance of $M$ (e.g., Tehran) being compared to every instance of $O$ (e.g., Karaj river), for every $O_i$ in $R$ (e.g. river) (Bogorny et al., 2005). An important contribution to this field is the Weka-GIS plug-in by Bogorny et al., for the classic data mining tool Weka. The user selects a database and a set of features. It calculates the distance and topologic relations between them, and Weka uses the resulting ARFF file to perform data mining. This is done regardless of any prior knowledge and provides automation in two granular level for the resulting file namely feature type granularity level and feature instance granularity level (Bogorny et al., 2005).

For mining spatial data using GP and top, a niche genetic programming algorithm named DMGeo has shown good results in classifying geographic data which also takes into account the topological relationships (de Arruda Pereira et al., 2010). As discussed, GP is more accurate in classification than other well-known methods. But the most challenging part is the time-consuming training phase. Nevertheless, in geographic data classification, large amounts of data are available. Many tasks including decision making in Geographic Information Systems (GIS) require faster knowledge discovery without losing accuracy. Many efforts have been implemented which require huge preprocess and execution time.

In this paper, to use GGeo for large databases and minimize the feature number which leads to better results, feature type granularity level is used (Bogorny et al., 2006). As discussed earlier, these crisp methods dismiss lots of knowledge, and using a fuzzy RCC membership function could be a great improvement.

In previous works, spatial relationships are calculated by database engine tools like PostgreSQL with PostGIS which only support crisp topological relations that assume clear boundaries for geographic features. However, most real-world geographic have imprecise and fuzzy boundaries that make crisp, non-fuzzy RCC less accurate. Our proposed architecture is based on a richer model of RCC namely fuzzy RCC which allows fuzzy boundaries for regions and will be introduced in section 3.

## 3. The proposed method

In this section, the proposed GGeo-T and GGeo-P architectures are presented after an overview of some important underlying concepts.

### 3.1. Evolutionary algorithms

Evolutionary Algorithms (EA) utilize stochastic searches as inspired from the natural evolution process. The EA methods often share following rules:

1- EAs work with a *population* of individuals (candidate solutions) rather than a single candidate solution.

2- EAs use a biased *fitness function* to evaluate the quality of an individual. The higher the score, the higher the inheritance chance of its "genetic material".

3- EAs create *new individuals* using probabilistic operators, which are usually crossover and mutation. The crossover operation exchanges genetic material between two or more individuals and mutation updates a part of the individual's genetic material.

The key concept in designing an EA is about choosing a suitable population representation and genetic operators. These parameters are biased and should be chosen according to the problem. The Genetic Programming (GP) is a newer variant of GA that uses a hierarchical tree for representation and crossover (lower probability than GA) or mutation (higher probability than GA). An individual (candidate solution) is represented as a *tree* with its internal nodes as "functions". The *leaves* in the tree are the variables or constants of the problem at hand. So, the main difference between GP and GA is that in GP an individual can take "functions" or "operators" as variable values, unlike GA which can just have normal values. So each individual in GP is called a "program" which is a recipe for solving a particular type of problems.

The function set for GP contains a collection of functions that different values of the variable can be assigned to them. This set should satisfy two conditions, *sufficiency* and *closure*. Sufficiency is defined as having the ability to represent a solution to the target problem. Closure refers to the fact that other functions must accept the output of a function as their input. Since GP usually works with different types of data, the c*losure* property is hard to satisfy in practice. A solution to this problem is to score non-consistent trees with the least fitness value. This is proven not to be efficient. Another option is converting the outputs of different data types. It works unless there is no natural conversion between those types. In fact, determining the function set is a trade-off between the expressivity power of the solutions and the cost of search in a large space. It has been confirmed that the set of {+, -, ?, *OR*, *AND*, *XOR*} satisfies both conditions.

An important aspect of evolutionary algorithms is their robustness. It means that when a problem is modeled, the algorithm can search all the available search space to find the best solution. In this regard, the MOSES (Meta-Optimizing Semantic Evolutionary Search) algorithm is developed as a new approach to program evolution which fits into the larger scheme of OpenCog framework. The framework is based on representation-building and probabilistic modeling. The MOSES algorithm has been applied to many application domains such as biology and text mining. It uses Lisp-like mini programs and cloud be used in a more limited aspect, as a supervised classification algorithm. A detailed description of the MOSES algorithm is presented in the following section.

**3.2. The MOSES automated program learning**

MOSES performs a supervised learning and uses a scoring function or training data as the input. MOSES sends a "*Combo program*" to output which estimates the scoring function. MOSES creates a population of programs and then searches the neighbors and "*mutated*" programs and calculates their fitness. After a certain number of iterations, the best program is represented as the output. In particular, MOSES constructs a population of "*demes*". Each deme is a program with many tunable parameters. These parameters are called "*knobs*". So finding the best program is not only about finding the best deme, but also includes choosing the best deme with the best knob settings.

MOSES chooses a deme and performs random mutations on it. This is done by inserting new knobs in random places. The best knob settings for each mutated deme is chosen using the best optimization algorithms like hill climbing, simulated annealing or by estimation of distribution algorithms such as the Bayesian Optimization. After this step, the fitness of the resulting deme is compared to the dominated (original) one. If the new program (deme) is more fit, the old dominated deme will be removed and the new one is set as dominated and this process will repeat.

All evolutionary algorithms tend to construct huge, bloated and convoluted code. To eliminate such codes, MOSES performs a reduction in each step to convert the program to a normal form. This normalization is based on the "*Elegant Normal Form*" by Holman (Holman, 1990). This is much more compact than the Boolean disjunctive normal form. So, normalization removes the redundant terms and makes the code more readable and quicker to execute. MOSES usually outperforms standard genetic programming due to its more efficient optimization and normalization. The OpenCog framework also includes a feature selection or dimensional reduction program. In machine learning problems, there are often a lot of variables that simply will not contribute to the final result. Eliminating these at the outset will improve the speed and accuracy of the machine learning system, so it is used in MOSES to improve the results (2014a).

### 3.3. The need for preprocessing along with MOSES

As described in section 3.1, determining the function set is crucial in GP. The function set has a high impact on the population number and the length of trees. In the case of using GP in classifying many records with many attributes, a large population and long individual length occur. This leads to a high computation load. On the other hand, the geographical functions for expressing the topological and geometric relationships between spatial data are of a high computational order. Adding these geographic functions to the sophisticated function set will result in a significant amount of calculation time and many repetitions. Moreover, satisfying the closure property requires a lot more computations.

A solution to these problems can be attained by an architecture which preprocesses geographical relations with as low impact on expressive power as possible. This will highly affect the time and memory requirements. Considering geographical relations calculated in preprocess phase as a sub-tree invisible to GP, these problems could be handled in a transparent manner.

An illustration is shown in Fig. 1. As can be seen, Fig 1a shows a chromosome (a MOSES individual) when the function *Overlaps* is in the function set. If the calculation of the marked part is done in the preprocess phase and then the result is added as a new attribute *Overlaps_Highway* to the data, the calculations will be invisible to GP (Fig. 1c). A single node is then added instead of the marked subtree of Fig. 1b, which makes the search faster and smaller, without affecting the accuracy of the method.

On the other hand, this approach can inspect the output values of functions independent of GP, and handle inconsistent data and functions. There is no need to employ any inconsistent sub-trees preventing method. In other words, GP is not able to apply geographic functions on non-geographical data and vice versa when creating inconsistent trees.

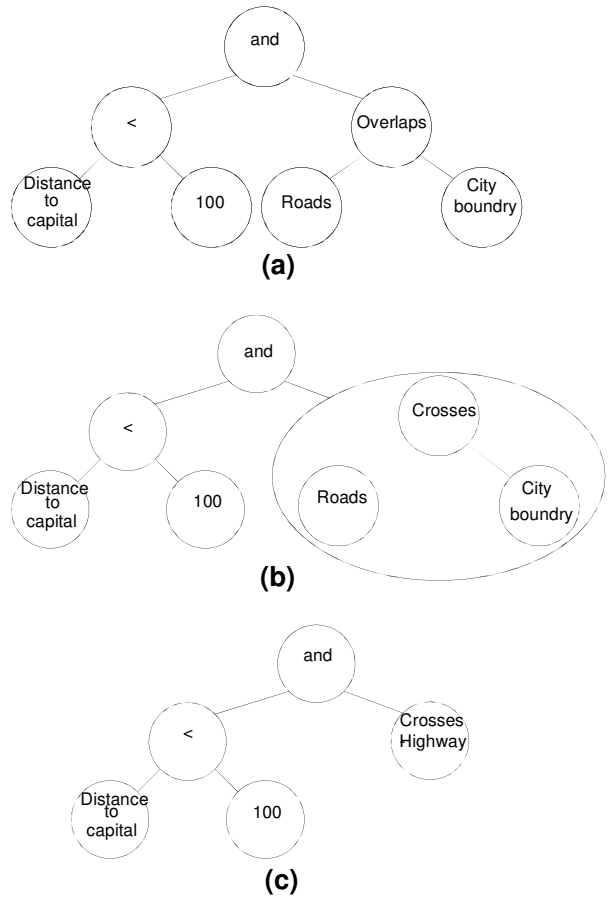

Fig 1. Represented data in GP

### 3.4. Spatial relationships

Topological relations are a particular class of spatial relations that are defined between two spatial or geographical entities. Other classes of spatial relationships are *distance relations* that set the order of objects. A distance relationship is a positive real number and can be also expressed with less precision with phrases like "*Close to*" and "*Far from*". *Directional relations* represent the relative position of an object to the other and needs the space to be directed. So a reference system and axis is necessary (e.g., for three-dimensional reasoning a system with relations "right of", "left of", "over", "under", "in front of", "back of", is required.

The boolean topological relations between two objects are based on the concepts of intersection, internality, and externality. A well-known formalism for topological relationships is RCC in which all the relations between entities are obtained from a connection predicate. Presented by Randal et al., the RCC model defines the topological relationships using a primary relation for connecting (shown as C) (Randell et al., 1992). The RCC model serves as a method for qualitative spatial representation and reasoning. The RCC model describes regions in Euclidean or topological space as well as their possible relationships. The RCC8 model consists of eight basic relations that are possible between two regions, four of them shown in Fig 2.

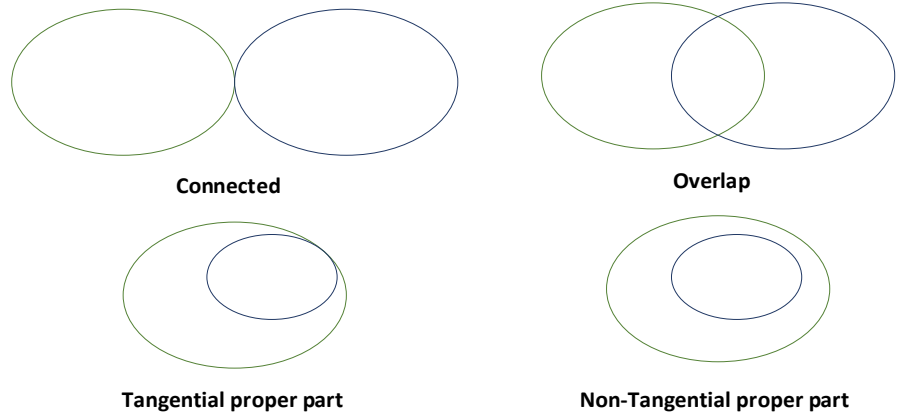

Fig 2.Some RCC Relations

From a practical view, the real-world regions that participate in an RCC relationship do not have clear and exact boundaries. Ambiguity is intrinsic property in geographical contexts. For example, when we refer to a forest, there is no sharp boundary to separate the forest region from its surrounding non-forest regions. Processing the sophisticated spatial relations and spatial reasoning are computationally intensive and handling this ambiguity makes the task even more challenging. On the other hand, simplifying the relationships by neglecting such imprecise boundaries will lead to loss of valuable knowledge that is required for making high-quality decisions.

To fill this gap, fuzzy RCC model has been proposed (Schockaert et al., 2008a), which is exploited in the present work.

The classic RCC evaluates a connectedness relationship like $C(x,y)$ as true, when $x$ and $y$ have a common point and as false when they don't. Using this primary "*Connected*" relation, other relationships are defined. On the other hand, the knowledge extracted from geographical relations is tacit knowledge. As discussed earlier, expressing geographic entities as fuzzy shapes and employing fuzzy methods on them allows handling vague regions better. A detailed definition of fuzzy topological relations is provided by Schockaert et al. (Schockaert et al., 2008a,Schockaert et al., 2008b,Schockaert et al., 2009). Their approach is briefly described hereafter an overview of the basic concepts of fuzzy sets. A fuzzy set in $\mathbb{X}$ is a mapping $A$ from $\mathbb{X}$ to $[0,1]$ which for each $x$ in $\mathbb{X}$, $A(x)$ is a degree of ambiguity. A fuzzy set $\boldsymbol{R}$ in $\mathbb{X} \times \mathbb{X}$ is a fuzzy relation in $\mathbb{X}$ and $\boldsymbol{R}(x, y)$ is a degree of relation satisfied between $\boldsymbol{x}, \boldsymbol{y}$. In order to define topological relations in the fuzzy RCC model, the primary fuzzy *Connection* relation is considered. Here $\boldsymbol{C}(\boldsymbol{u}, \boldsymbol{v})$ in $\mathbb{U}$ shows the degree (a number between 0, 1) to which the regions $\boldsymbol{u}, \boldsymbol{v}$ are connected. For generalizing a fuzzy relation, fuzzy t-norms are used. In this paper Łukasiewicz t-norm is used.

$$C_{(\alpha,\beta)}(A, B) = \sup_{p \in R^n} T(A(p), \sup_{p \in R^n} T(R_{(\alpha,\beta)}(p, q), B(q)))$$

$$O_{(\alpha,\beta)}(A, B) = overl(R_{(\alpha,\beta)} \downarrow\uparrow A, R_{(\alpha,\beta)} \downarrow\uparrow B)$$

Other fuzzy relations can be easily defined by using the connect relation. The implementation of fuzzy RCC relationships is a challenging task that leads to heavy computations. We will propose an approximation method for fuzzy RCC computation in the GGeo architecture.

**3.5. The proposed architecture – GGeo**

In this section, the GGeo architecture is presented as a novel hybrid data mining approach that utilizes a combination of spatial data preprocessing and MOSES automated program learning algorithm. It performs classification tasks on geographic databases in an efficient manner.

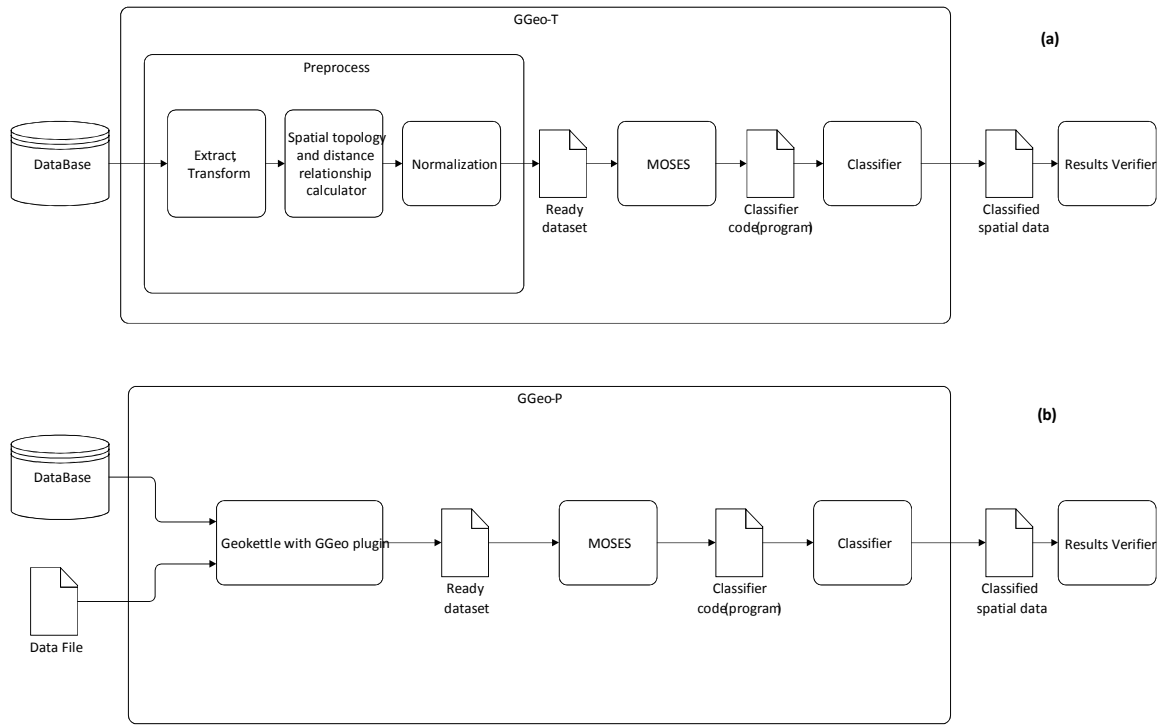

Fig 3.The proposed GGeo-T and GGeo-P architectures

GGeo unifies geographic constraints and operation at each instance. Two related architectures are proposed as can be seen in Fig 3 that are described as follows.

3.5.1. Inputs and outputs of GGeo

The inputs to the system are a collection of numerical, Boolean, spatial features. For example, in classifying cities, the numerical feature types are population, Gini index and the Boolean features are provincial capital city and geographic feature types like the city map, roads, and rivers which are in line and polygon form.

GGeo creates a different program for each class by learning from the training data (for example, a few cities). Then it can inspect the test data and classify it accordingly. So the outputs are classified data, and the generated classifying program (rules). The second architecture, GGeo-T, further improves this through a pre-process and supervised data mining (train-test) parts. In the pre-process part, after the initialization of the connection between GGeo and the geographic database, raw data is read from the database. These data contain spatial data tables which represent a geographic feature and its attributes. Attributes in this table are numerical, Boolean, nominal and spatial. The spatial attributes are geographic points, lines, and polygons. First the target feature (TF) and the relevant features whose spatial relations toward TF are important, are selected. Also, the particular relationship between the target feature type and the appropriate feature type is set.

$T_F$ = Target Feature

$F_s = \{F_i \mid 1<i<n\}$

$R_S = \{R_{Si} \mid 1<i<n, R_{Si} = \{R_{Sij} \mid 1<j<m\}\}$

In which n is the number of relevant features, m is the number of spatial relations for each relevant feature and $R_{Sij}$ is a spatial relation (topologic or distance) - between $T_f$ and $F_i$. Calculating topological relations is shown in Algorithm 1.

```
1.  Foreach instances of TF as t do begin
2.   For i=1 to n do begin
3.    Foreach instances of Fi as f do begin
4.     Join TF and f
5.     For j=1 to m do begin
6.      If not exists, Add a new attribute f+Rij to TF
7.       Value = Calculate f, Rij
8.        Add t, Value
9.     End
10.   End
11.  End
12. End
```

**Algorithm 1**.Topological relation calculus using low granulation Level

This granularity level is very small. A higher level should be used if we need to avoid adding too many attributes and making it useful for decision making. So, instead of calculating the spatial relation between an instance of the target feature and each instance of a related feature type and reporting it directly, they are all aggregated for that feature type and reported. For example, when calculating the distance between the cities and roads, instead of reporting and adding the new attributes as (*city_road1, city_road2,...*) they are aggregated, and a single attribute *city_road* is added to data. This method is used in Algorithm 2.

```
1. For each instance of TF as t do begin
2.  For i=1 to n do begin
3.   If not exists, Add a new attribute Fi+Rij to TF
4.    For each instance of Fi as f do begin
5.     Join TF and f
6.     For j=1 to m do begin
7.      NewValue = Calculate f, Rij
8.      Value = Aggregate Value, NewValue
9.      Add t, Value
10.    End
```

```
11.      End
12.     End
13.    End
```

**Algorithm 2.** Topological relation calculus using high granulation level

So with different geographical feature types as input, new numerical attributes are calculated which represent different spatial relations between those geographic types. A schematic view of this process is shown in Fig 4.

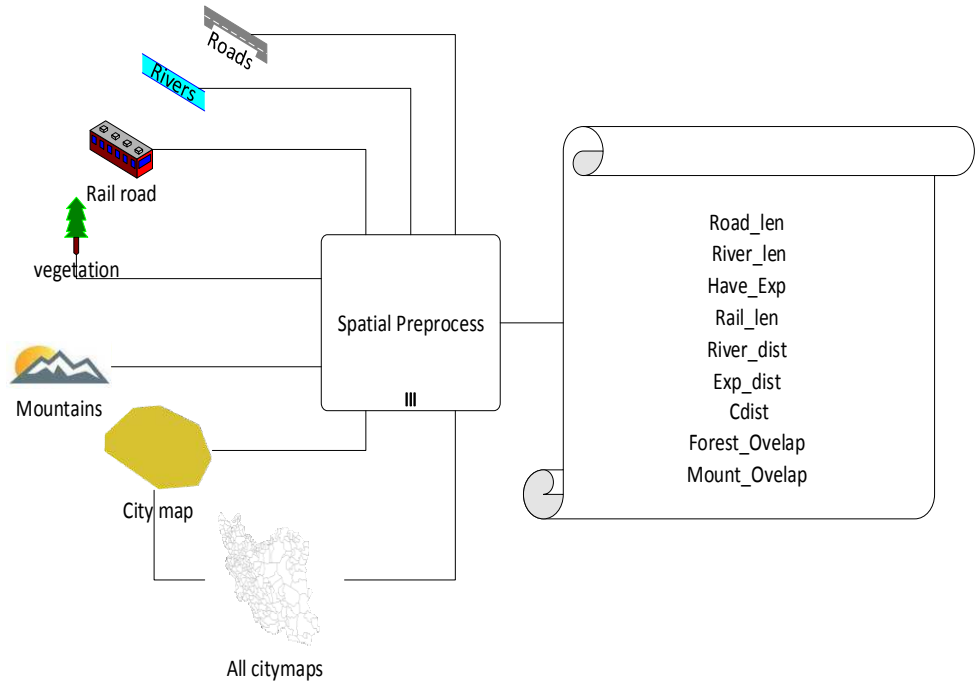

**Fig 4.** Inputs and outputs of spatial preprocess

3.5.2. Inputs and outputs of GGeo

Calculating the value of the spatial relations between different feature types is done with the help of fuzzy RCC.

```
1. indexing():
2. $grid = Grid($Scope, $n);
3.  Foreach featureType as $f do begin
4.   If isPoligon($f) then
5.    Foreach $f as $i do begin
6. Foreach instance in $grid as $g do begin
7.   MemberShip[$f][$i] = CalcMF(`Area`,$g,$i);
8.  End
9. End
10.     Elseif isLine($f) then
11.    Foreach $f as $i do begin
12.     Foreach instance in $grid as $g do begin
13.      MemberShip[$f][$i] =
                CalcMF(`length`,LenIn($g, $i));
```

```
14.     End
15.    End
16.     Elseif isPoint($f) then
17.   Foreach $f as $i do begin
18.    Foreach instance in $grid as $g do begin
19.     MemberShip[$f][$i] =
                    CalcMF(`count`,numPointsIn($g, $i));
20.    End
21.   End
22.    End
23.    End
24.  R($x,$y,$alpha, $betha):
25.    $d = distance($x,$y);
26.    If($d<=$alpha) Return 1;
27.    If($d>$alpha+$betha) Return 0;
28.    Return ($alpha+$betha-$d)/$betha
29.  Tw($a, $b):
30.    Return Max(0, $a+$b-1);
31.  It($a,$b,$precison=0.01):
32.    Return Min(1, 1-$a+$b);
33.  Overlap($F1, $F2):
34.    InfIt1 = InfIt2 = 1; SupTw1 = SupTw2 = 0;
35.    Foreach $grid as $x
36.     Foreach $grid as $y
37.      SupTw1=Max(SupTw1,Tw(R($x,$y),MemberShip[$F1]));
38.      SupTw2=Max(SupTw2,Tw(R($x,$y),MemberShip[$F2]));
39.     End
40.    End
41.    Foreach $grid as $x
42.     Foreach $grid as $y
43.      InfIt1=min(SupTw1,It(R($x,$y),SupTw1));
44.      InfIt2=min(SupTw2,It(R($x,$y),SupTw2));
45.     End
46.    End
47.    Return Tw(Inflt1,Inflt2);
```

**Algorithm 3.** Tiling and Fuzzy Overlap Calculation

As could be seen in Algorithm 1 and Algorithms2, reading the geometries of geographical features, different spatial relations should be calculated among them. This work contains distance relationship and fuzzy overlap degree. The geographic scope of the problem is divided and tiled to a $n \times n$ grid. to implement the spatial relationship preprocessing in the present work. In our case, the scope is the enclosing rectangle of the map of Iran that will be described in Section 4. We need to map the problem to fuzzy RCC, so each cell (tile) in the grid is supposed an $x$ in $\mathbb{X}$ and has different degrees of fuzzy membership in different geographical entities. Based on calculating the memberships and applying fuzzy RCC on them as in Algorithm 3, the fuzzy overlap functionality is implemented and used in the proposed hybrid architecture, GGeo.

The inputs of different problems in the spatial classification of cities are presented in Table 1. For multiline features like roads and rivers, the length of the feature contained in the target feature (city) is calculated. Considering a minimal width for the lines and change them to minimal polygons, the problem can be transformed to fuzzy overlap RCC. Tiling each scope and supposing each tile as a member in fuzzy geographical features sets, the overlap degree is calculated for each tile in the $n \times n$ map, as Algorithm 3.

Table 1. Typical problems and their transformation into fuzzy RCC

| Problem | Geometric problem | Transformed problem in Fuzzy RCC | Inputs |
|---|---|---|---|
| Total road length for city | Length of line segment in polygon | overlap | $L=\{l_1,l_2,\ldots,l_n\}$ $l_n=\{(x_{n1},y_{n1}),(x_{n2},y_{n2}),\ldots,(x_{nm},y_{nm})\}$ $P=\{(x_1,y_1),(x_2,y_2),\ldots,(x_p,y_p),(x_1,y_1)\}$ |
| Total river length for city | | | |
| Total rail road length for city | | | |
| Distance to river | Distance between line segment from polygon | - | - |
| Distance to Expressway | | | |
| Distance to province capital | Distance between two polygons | | $P_1=\{(x_{11},y_{11}),(x_{12},y_{12}),\ldots,(x_{1p},y_{1p}),(x_{11},y_{11})\}$ $P_2=\{(x_{21},y_{21}),(x_{22},y_{22}),\ldots,(x_{2q},y_{2q}),(x_{21},y_{21})\}$ |
| Overlap on forest | Area of the intersection of two polygons | overlap | |
| Overlap on mountains | | | |

As geographic data have many different formats and handling all these formats will require translation between them, GGeo should also be deployed as a plugin for a spatial ETL (Extract-Transform-Load) tool, GeoKettle.

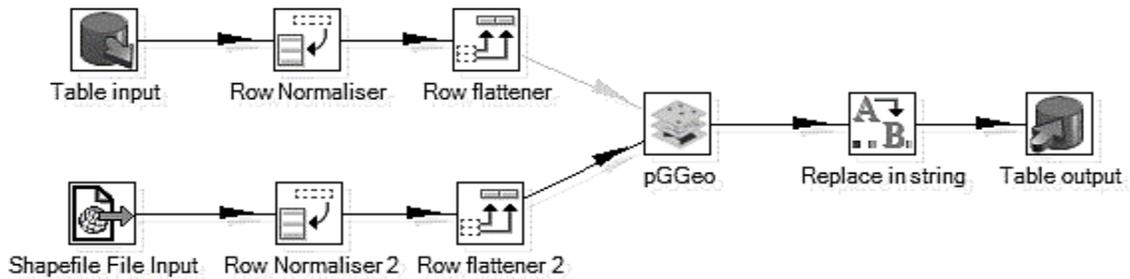

**Fig 5**. GGeo as a plugin for Pentaho (GeoKettle)

GeoKettle automatically extracts data from different sources and performs cleaning, manipulation, standard transformation and finally loads them in a conventional DBMS, GIS file or a spatial web service. For this to happen, our second architecture GGeo-P is proposed, implemented in four java files as a standard GeoKettle plugin.

## 4. Experimental evaluation

In this section, the proposed GGeo architecture is applied to real-world data, and the results are represented as a proof of its efficiency.

### 4.1. Raw dataset

The spatial data sets used in this research are in the WGS84 system. The geographical points are shown with latitude and longitude.

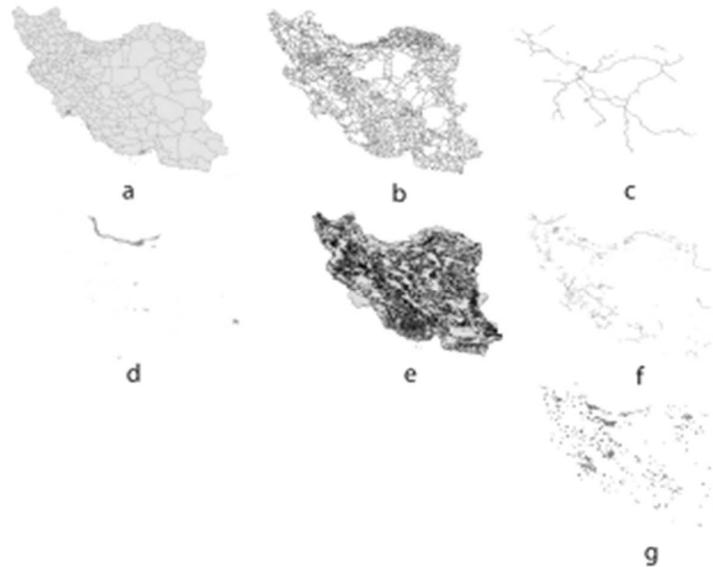

**Fig 6.** Geographical features for the experiments

Each of these properties is shown using decimal floating points with double precision. *Cities of Iran* data from GADM are geographic polygons which are sets of points with the same start and end points. In each city general attributes like population and last update, the date is also provided. *Iranian roads* are extracted from OpenStreetMap-OSM in the form of geographical lines (sets of ordered points). Other Iranian map data are from MapCrusine website.

After data gathering, the geometric data in WKT format were extracted. The entire map scope was tiled with vertical and horizontal grid lines to calculate the spatial relationships of the features.

Considering each cell as a member of the *universal set* and defining a fuzzy membership function for each geographic feature, the fuzzy RCC relationship between two features is converted to the fuzzy RCC relation between two fuzzy sets(As described in Section 3.4).

Due to the low precision in geometric calculations with adjacent data values, to apply the spatial processing, a new discrete coordination system is adopted as follows. In the tiled map ($n \times n$), each tile is divided to a $m \times m$ area and each part is called a pixel. So the whole map is represented by a ($m*n \times m*n$) matrix.

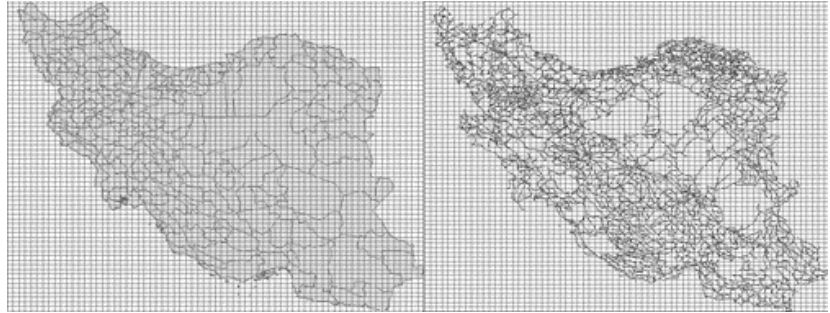

**Fig 7.** Two samples of the Iran tiled maps

Parameter *m* is considered as a *level* of granulation. For each feature type, it is presented in the new coordinate system and it is drawn in a bitmap matrix. So for each grid cell, the number of pixels painted, show the approximate area of cell covered by the geographic feature. This approximate number is used as the input of a fuzzy membership function (Fig 8), and returns the tile's fuzzy membership in a special feature (as done on Algorithm 3, indexing procedure, lines 1-23).

Therefore, after the fuzzy overlap and other desired spatial relations are calculated for each instance of the target feature (as seen in algorithm 2,3), they are inserted as new attributes for the instance. So, to formally address the city classification problem, the inputs are the cities maps as the target feature type, rivers, railroads, mountain and plains, vegetation and urban areas as the relevant feature types and the required spatial relations are fuzzy RCC overlap and distance relationship.

$T_F$ = city

$F_S$ = {river, road, rail, mount, veg, urban}

$R_S$ = {{distance, length}, {distance, length}, {distance, length}, {overlap}, {overlap}, {overlap}}

So according to Algorithm 2 the new attributes in the preprocess are calculated as $F_s \times R$.

$F_S \times R$ = {`river_dist`, `river_len`, `road_dist`, `rail_len`, `rail_dist`, `rail_len`, `mount_overlap`, `veg_overlap`, `urban_overlap`}

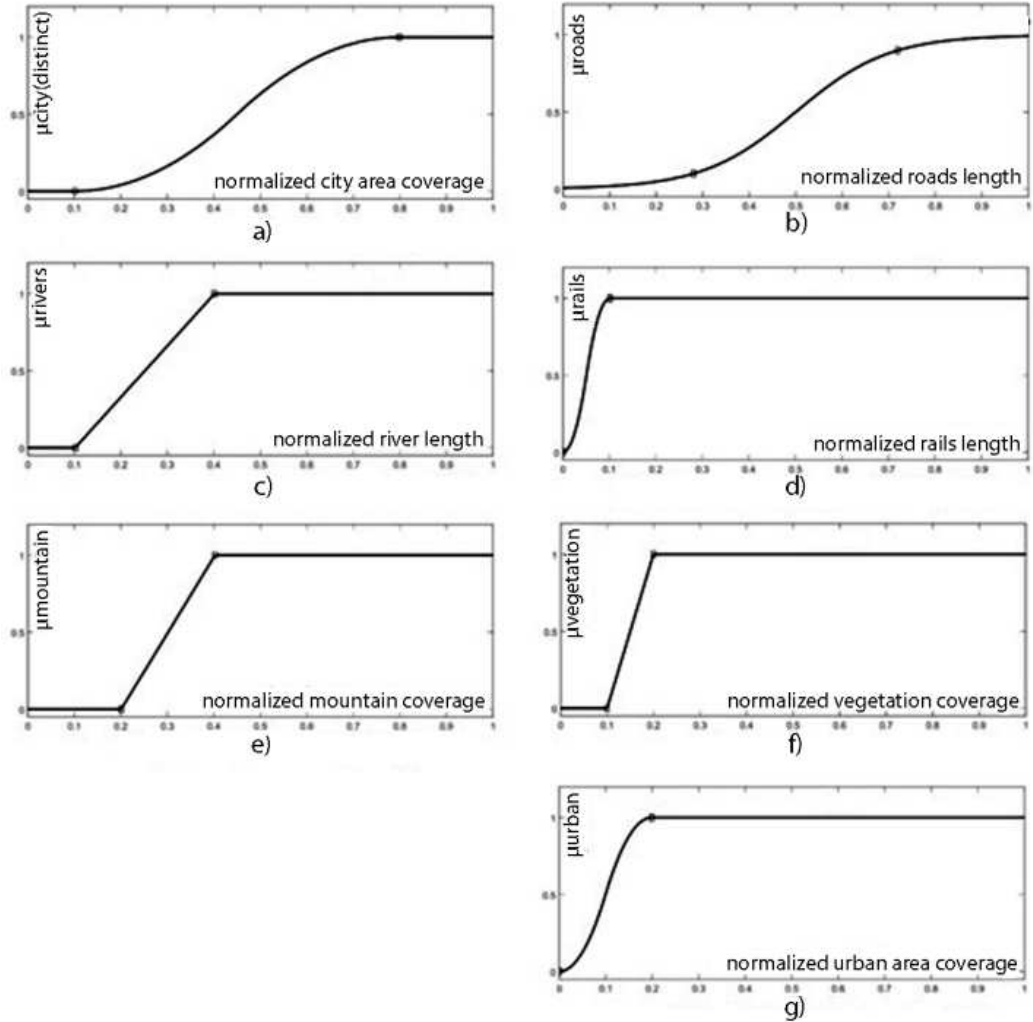

**Fig 8.** Fuzzy membership functions for different feature types

These new attribute values are the calculated fuzzy membership degrees for the target feature type instance and the relevant feature types. Two additional attributes *Have_Exp* and *Cdist* (containing expressway and distance to the province capital), defining *ExpressWay* ⊆ *Road* and *Capital* ⊆ *City* were also used.

After normalization of previous and new attributes, MOSES algorithm is applied on the data. As pre-process is very time consuming, both calculating memberships and fuzzy RCC relationships, are calculated in parallel. The performance of calculating these membership values for Iranian cities case in normal and parallel is shown in Fig 9. Calculating fuzzy overlap between geographic feature types in Iranian cities case is shown in Fig 10.

In the city development dataset, three classes are defined. A total of 264 cities are labeled high class(A), 296 are labeled medium (B), and 292 are labeled low (C). In this dataset, 22 numeric attributes such as the number of schools, gini coefficient and also spatial geographical attributes like city maps(as polygon) and railroad and highways(as lines) are present. Soy Aptitude data set contains the data of two types of soil, which one is suitable(A) and the other is not suitable(B) for growing soya beans. 562 of them are appropriate, and 290 of them are not. Balanced Soy aptitude is a selection of the prior by Priera et al.(de Arruda Pereira et al., 2010) which 290 are of class A and 290 are of B.

Iran cities data set is generated by the authors under the supervision of urban planning experts and contains spatial and non-spatial data of 268 Iranian cities like their population, the number of hotels and educational index. 53 of them are labeled High (A), 128 of them are labeled Medium (B) and 87 of them are labeled Low(C).

The performance of the proposed method GGeo is analysed by applying it in 6 classification problems. Two datasets(Heart and Wine databases) containing non-spatial data are available in the UCI Machine Learning Repository(2014b) and are used only to evaluate MOSES. Heart dataset contains 270 instances, 120 of which are labelled as sick (B) and 150 as healthy (A). Wine dataset has three different classes, 59 of which are A, 71 are B and 48 are C. City development and Soy Aptitude dataset are provided by Geominas which contain numeric and spatial attributes.

To use traditional data mining approaches on data, Weka-GDPM was used and these data sets are labeled non-spatial in the result table (Table 2). The genetic programming algorithm in the tests (DMGeo) is iterated to 200 generations, with 90% probability of crossover and 2% probability of mutation. It uses crisp Boolean topological relationships implemented in PostgreSQL.

The results shown in Table 2 show that fuzzy spatial preprocessing importing more spatial knowledge in classification, is more efficient. Comparing GGeo on City Development dataset and on non-spatial City Development dataset, this is brightly shown. Regarding the results, it can be said that GGeo works better in the geographical contexts and is more stable than DMGeo.

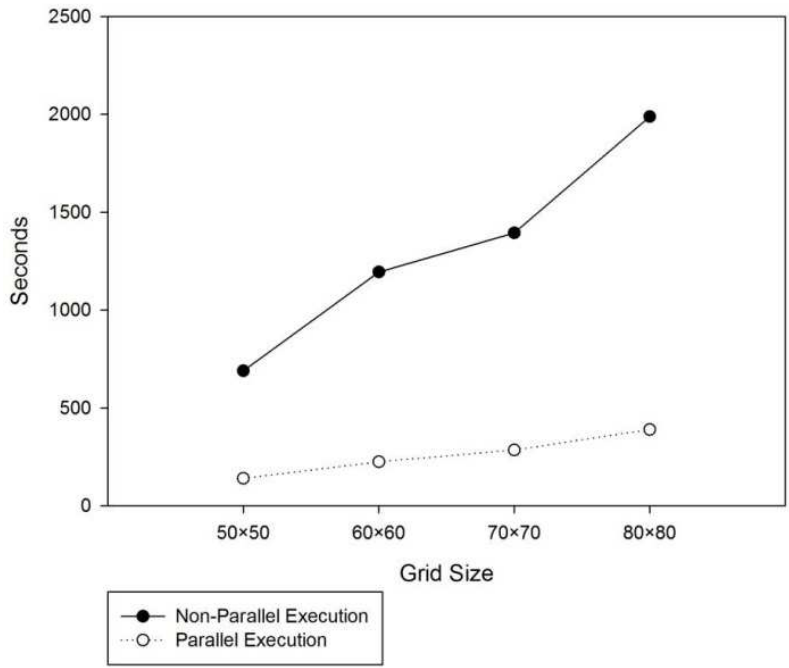

**Fig 9.** Comparing normal and parallel calculation of Fuzzy membership of each pixel

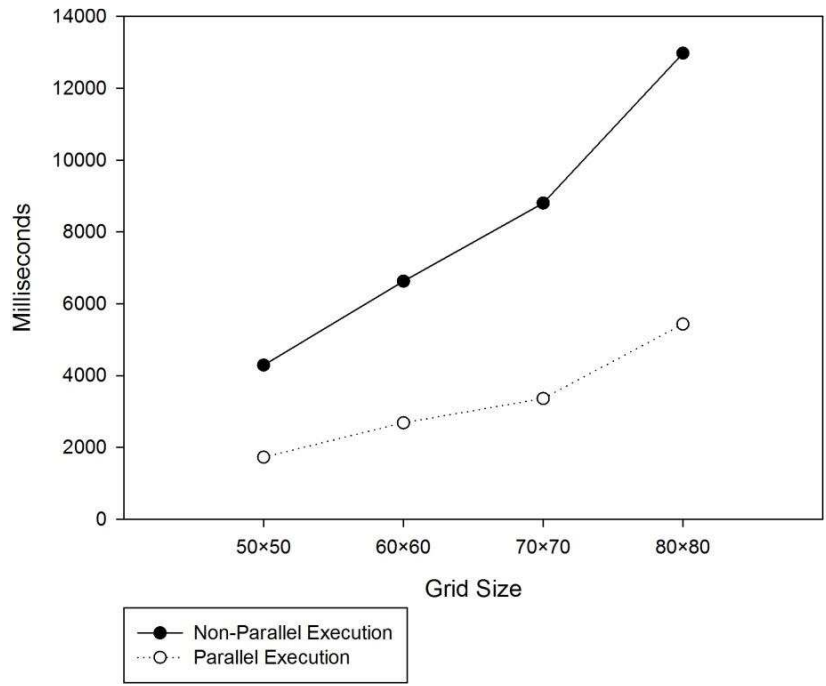

**Fig 10.** Comparing normal and parallel calculation of Fuzzy Overlap degree for each tile

**Table 2.** Results of different data mining approaches on datasets

| DataSets | J48 | RBF | SVM | DMGeo | GGeo |
|---|---|---|---|---|---|

| Dataset | | | | | | | | | | | | | | | | |
|---|---|---|---|---|---|---|---|---|---|---|---|---|---|---|---|---|
| Wine | A | 96% | | | 99% | | | 98% | | | 89% | | | 91% | | |
| | B | 85% | | 90% | 96% | | 97% | 100% | | 99% | 86% | | 92% | 92% | | 93% |
| | C | 90% | | | 95% | | | 98% | | | 95% | | | 96% | | |
| σ | | 5.52 | | | 2 | | | 2.63 | | | 5.8 | | | 2.1 | | |
| Heart | A | 77% | | | 75% | | | 80% | | | 85% | | | 73% | | |
| | B | 79% | | 78% | 83% | | 79% | 87% | | 83% | 73% | | 79% | 70% | | 72% |
| σ | | 1.8 | | | 2.51 | | | 0.7 | | | 10.17 | | | 3.15 | | |

| | | B | N | P | B | N | P | B | N | P | B | N | P | B | N | P |
|---|---|---|---|---|---|---|---|---|---|---|---|---|---|---|---|---|
| Soy | A | 62% | 68% | | 65% | 60% | | 73% | 75% | | 93% | 90% | | 99% | 88% | |
| | B | 22% | 67% | | 11% | 58% | | 74% | 70% | | 59% | 70% | | 45% | 79% | |
| σ | | 31 | 0.1 | | 3.5 | 3.1 | | 1.1 | 2.8 | | 27 | 15 | | 2.5 | 2.2 | |

| Dataset | | | | | | | | | | | | | | | | |
|---|---|---|---|---|---|---|---|---|---|---|---|---|---|---|---|---|
| CD-non spatial | A | 78% | | | 73% | | | 77% | | | 62% | | | 65% | | |
| | B | 50% | | 63% | 51% | | 57% | 43% | | 56% | 52% | | 61% | 75% | | 63% |
| | C | 62% | | | 50% | | | 49% | | | 68% | | | 48% | | |
| σ | | 12.58 | | | 17.86 | | | 19.65 | | | 6.89 | | | 5.51 | | |
| City | A | ✕ | | | ✕ | | | ✕ | | | 80% | | | 81% | | |
| | B | ✕ | | ✕ | ✕ | | ✕ | ✕ | | ✕ | 73% | | 79% | 80% | | 86% |
| | C | ✕ | | | ✕ | | | ✕ | | | 83% | | | 97% | | |
| σ | | ✕ | | | ✕ | | | ✕ | | | 3.89 | | | 1.58 | | |
| iran cities | A | ✕ | | | ✕ | | | ✕ | | | 82% | | | 87% | | |
| | B | ✕ | | ✕ | ✕ | | ✕ | ✕ | | ✕ | 74% | | ۷۹% | 78% | | ۸۳% |
| | C | ✕ | | | ✕ | | | ✕ | | | 80% | | | 85% | | |
| σ | | ✕ | | | ✕ | | | ✕ | | | 4.57 | | | 0.89 | | |

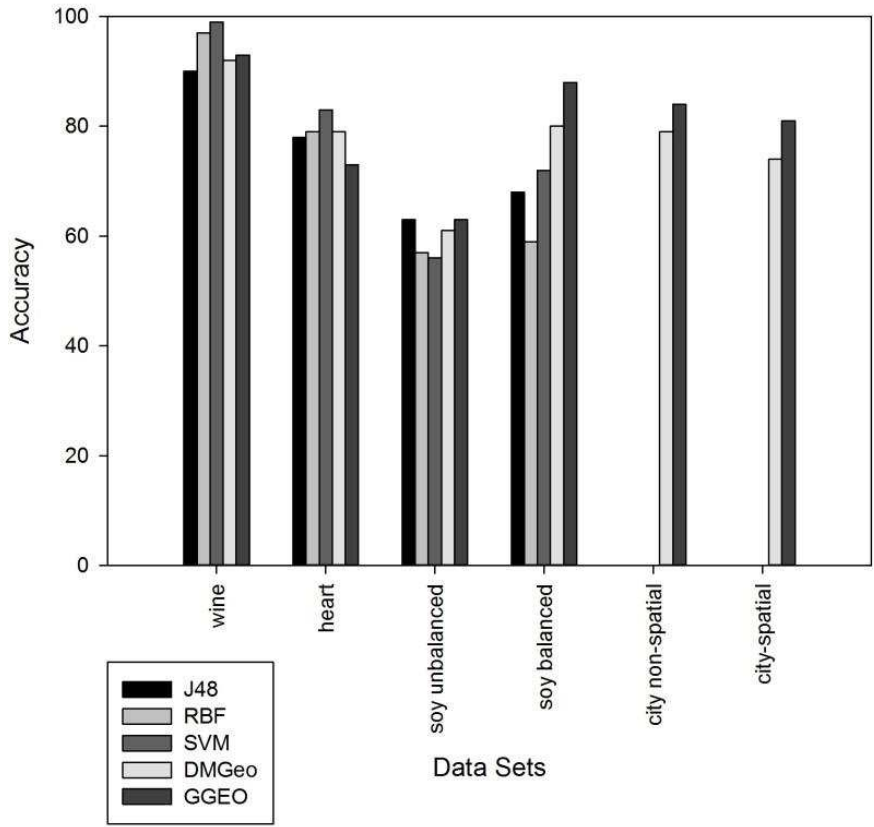

**Fig 11.** Accuracy in different datasets

To illustrate and compare GGeo and other methods, the accuracy and standard deviation of the results are calculated and shown in Fig.11 and Fig.12.

To compare GGeo and other approaches in noisy environments, two noisy datasets were generated. This was performed by manipulating the dataset initial spatial values. Before the pre-processing phase, spatial attributes are changed randomly. Consider $p$ as the probability of an instance containing error, and $q$ as the percentage of error. The noisy datasets using these parameters are fed into different approaches and compared (Table 3).

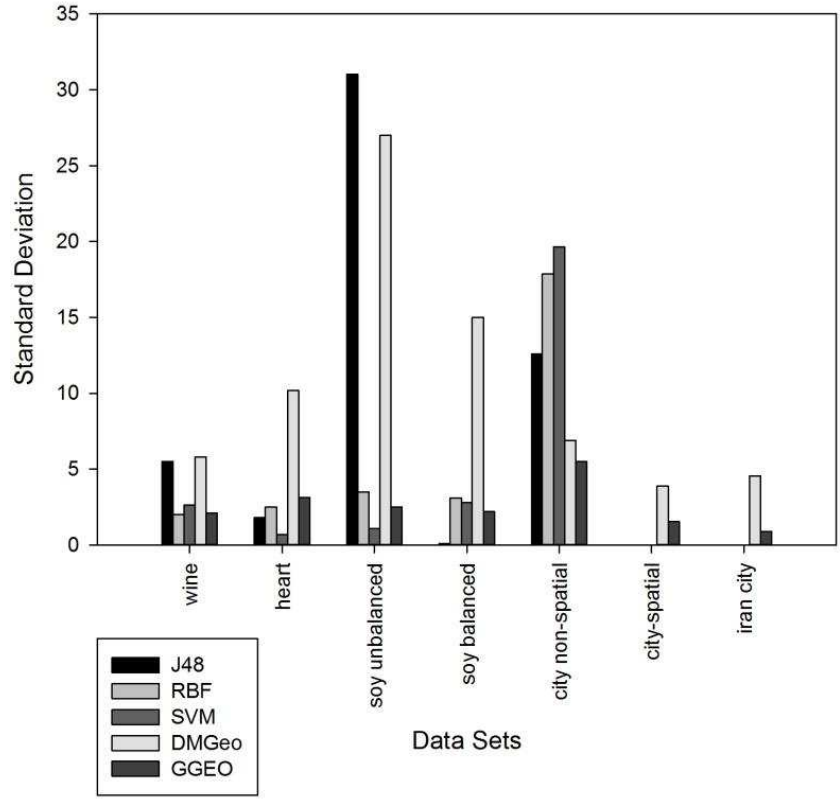

Fig 12. Standard deviation in datasets

By comparing the results, it could be observed that GGeo benefiting from fuzzy pre-processing and MOSES is much more resistant to noisy data. It is done using the value and the regression in Fig 13 and Fig 14.

Table 3. Results for the noisy datasets

| DataSet | parameters | GGeo | GP (DMGeo) | J48 (non-spatial) | SVM (non-spatial) |
|---|---|---|---|---|---|
| Iran Cities | p=0.1 q=0.1 | 85 | 72 | 56 | 48 |
| Iran Cities | p=0.1 q=0.2 | 78 | 67 | 53 | 45 |
| Iran Cities | p=0.2 q=0.1 | 85 | 70 | 47 | 45 |
| Iran Cities | p=0.2 q=0.2 | 76 | 62 | 45 | 44 |
| City Development | p=0.1 q=0.1 | 80 | 74 | 60 | 52 |
| City Development | p=0.1 q=0.2 | 76 | 74 | 58 | 50 |
| City Development | p=0.2 q=0.1 | 78 | 62 | 52 | 53 |
| City Development | p=0.2 q=0.2 | 70 | 68 | 45 | 47 |

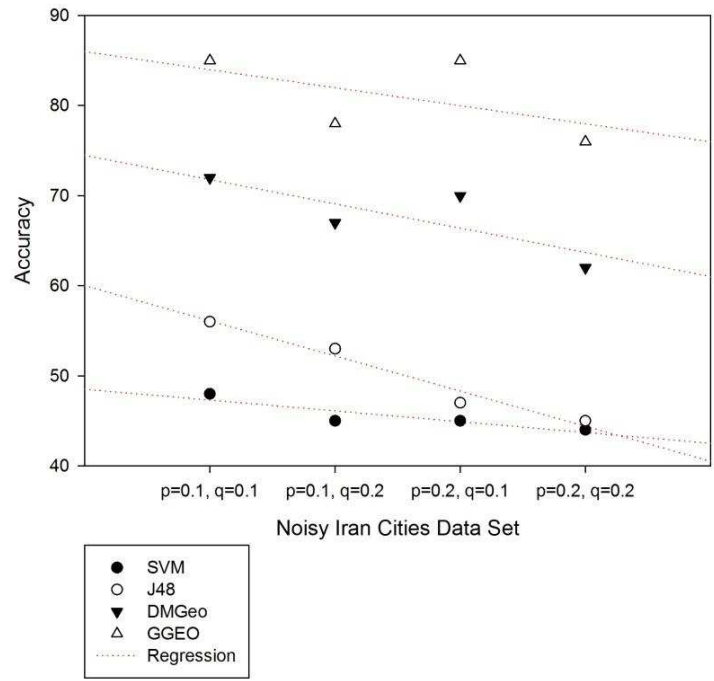

**Fig 13.** Comparing different approaches in noisy data injected in Iran City data set

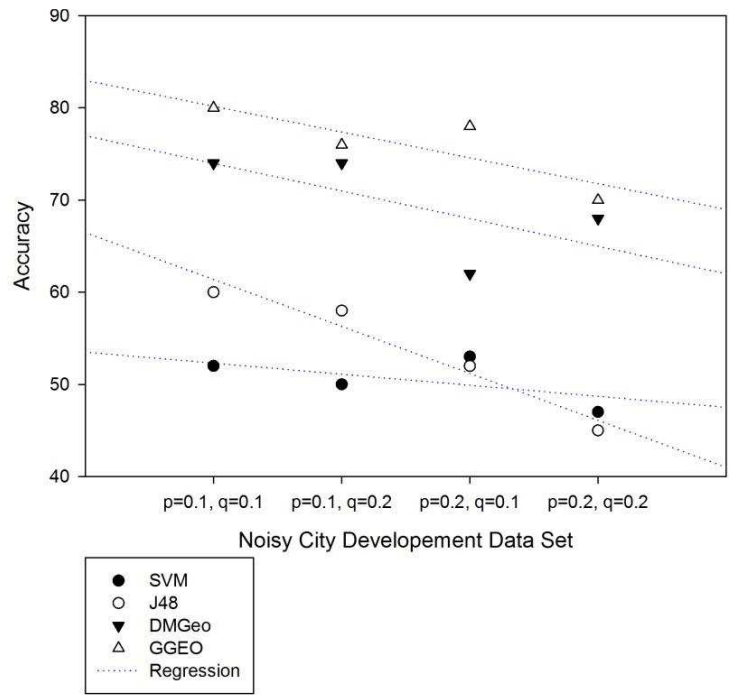

**Fig 14.** Comparing different approaches in noisy data injected in City Development data set

## 5. Conclusion and future work

Many emerging decision-making scenarios rely on geographic data that incorporates classification and other data mining tasks. Geographic data classifiers partition a region into categories based on the similarities between environmental attributes. Given the inherent uncertainty and incompleteness in spatial data, the fuzzy spatial relations are a vast knowledge source, often ignored in spatial data mining. This paper proposed GGeo as a novel hybrid data mining approach for classifying spatial features such as cities in geographic databases where spatial and non-spatial data are present. Iranian cities were used as a case study to illustrate the approach. The extensive evaluation showed that using fuzzy topology relations can significantly increase the quality of knowledge discovery in geographic databases.

Moreover, the GGeo architecture benefiting from the evolutionary MOSES algorithm is competitive and robust and has an excellent noise tolerance. A pre-processing approach that was proposed as a plugin for Geokettle further improved the algorithm. Future work may include using the association rules extracted by GGeo in a strategic decision-making problem. More spatial relationships could be also integrated into the proposed algorithm to improve the quality of decisions.

**Acknowledgements**

We would like to thank the OpenCog Project members for helping in the development of the initial idea and introducing MOSES to us. We also thank Marconi de Arruda Pereira for the datasets and DMGeo source code.